\documentclass[10pt,twocolumn,letterpaper]{article}

\usepackage{iccv}
\usepackage{times}
\usepackage{epsfig}
\usepackage{graphicx}
\usepackage{amsmath}
\usepackage{amssymb}
\usepackage{multirow}
\usepackage{booktabs}
\usepackage{xcolor, color, colortbl}
\usepackage{array}
\usepackage{tabularx}
\usepackage{pifont}
\usepackage{arydshln}
\usepackage{makecell}
\usepackage{enumitem}
\newcolumntype{H}{>{\setbox0=\hbox\bgroup}c<{\egroup}@{}}

\usepackage[pagebackref=true,breaklinks=true,letterpaper=true,colorlinks,bookmarks=false]{hyperref}

\newcommand{\myparagraph}[1]{\vspace{2pt}\noindent{\bf{#1}}}
\newcommand{\cmark}{\ding{51}}

\definecolor{colorwheel}{rgb}{0.0, 0.5, 1.0}

\iccvfinalcopy 

\def\iccvPaperID{5759} 

\ificcvfinal\fi

\begin{document}

\title{Denoising and Selecting Pseudo-Heatmaps for \\Semi-Supervised Human Pose Estimation}

\author{Zhuoran Yu\thanks{These authors contributed equally to this work.} \thanks{Currently at The University of Wisconsin–Madison. Work conducted during an internship with AWS AI Labs.}, $\;$ Manchen Wang\footnotemark[1] \thanks{Corresponding author.}, $\;$ Yanbei Chen, $\;$ Paolo Favaro, $\;$ Davide Modolo\\ 
AWS AI Labs \\
{\tt\small zhuoran.yu@wisc.com, $\;$ \{manchenw,yanbec,pffavaro,dmodolo\}@amazon.com}}

\maketitle
\ificcvfinal\thispagestyle{empty}\fi

\begin{abstract}
    We propose a new semi-supervised learning design for human pose estimation that revisits the popular dual-student framework and enhances it two ways. First, we introduce a denoising scheme to generate reliable pseudo-heatmaps as targets for learning from unlabeled data. This uses multi-view augmentations and a threshold-and-refine procedure to produce a pool of pseudo-heatmaps. Second, we select the learning targets from these pseudo-heatmaps guided by the estimated cross-student uncertainty. We evaluate our proposed method on multiple evaluation setups on the COCO benchmark. Our results show that our model outperforms previous state-of-the-art semi-supervised pose estimators, especially in extreme low-data regime. For example with only 0.5K labeled images our method is capable of surpassing the best competitor by 7.22 mAP ($+25\%$ absolute improvement). We  also demonstrate that our model can learn effectively from unlabeled data in the wild to further boost its generalization and performance. 
    \vspace{-2mm}
\end{abstract}

\newcommand{\ProposedMethodName}{{{{Ours}}}}
\newcommand\hl[1]{%
  \bgroup
  \hskip0pt\color{red!80!black}%
  #1%
  \egroup
}

\section{Introduction}
\label{sec:intro}

Deep neural networks have achieved remarkable success in the past decade thanks to the availability of large-scale annotated
datasets such as ImageNet~\cite{deng2009imagenet} for image classification and MS COCO~\cite{lin2014microsoft} for object detection and human pose estimation. However, obtaining a fully-labeled dataset at scale is extremely expensive in terms of both time and budget. 
In the case of human pose estimation, which is of interest in this work, such a dataset is even more challenging to build as it requires fine-grained and accurate annotations.

\begin{figure}
    \centering
    \includegraphics[width=0.65\linewidth]{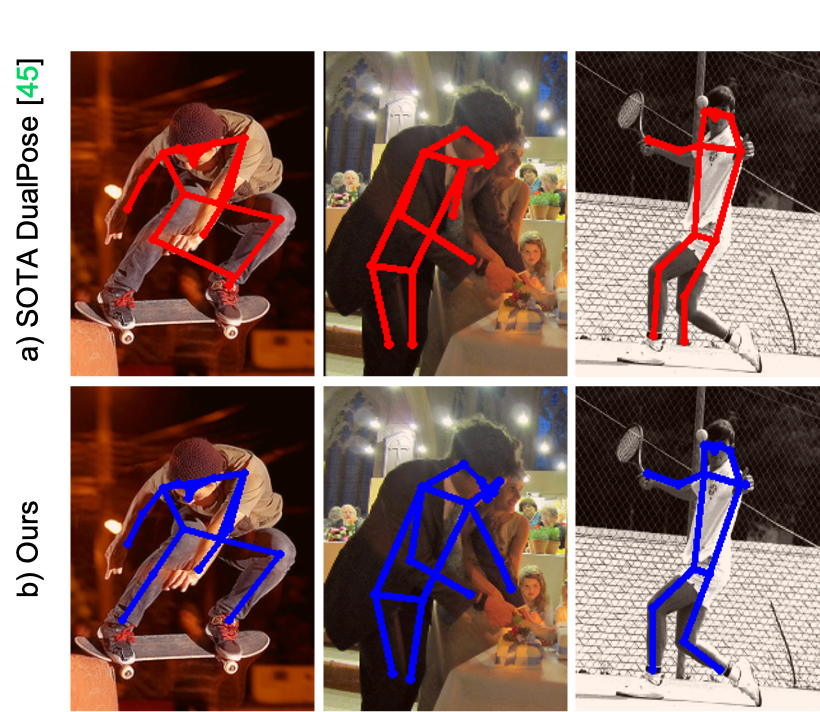}
    \caption{ \small \it 
    Given person images in different challenging poses, an existing semi-supervised pose estimator tends to produce noisy targets on unlabeled images (top row). By contrast, we propose to generate learning targets of higher quality (bottom row) by ensembling the heatmaps from multiple augmented views and selecting more reliable targets guided by uncertainty.
    \vspace{-4mm}
    }
    \label{fig:teaser}
\end{figure}

To mitigate the annotation cost in human pose estimation, we explore the use of semi-supervised learning (SSL)~\cite{zhu2005semi, zhu2009introduction, chapelle2009semi}, a field of research that has received increasing attention in the scientific community. Recent SSL techniques make it possible to build high-performing models by training them simultaneously on small labeled and large unlabeled datasets for image classification~\cite{lee2013pseudo, tarvainen2017mean, laine2016temporal, rizve2021defense, xie2020self, sohn2020fixmatch} and object detection~\cite{jeong2019consistency, sohn2020simple, zhou2021instant, liu2021unbiased, liu2022unbiased, jeong2019consistency}.

A key mechanism in recent SSL methods is the use of so-called \emph{pseudo-labels}, i.e., the predictions of a model during its training, as a replacement for the ground truth targets. 
Much of the success of SSL methods is related to the quality of the pseudo-labels, as errors can lead to confirmation bias \cite{arazo2020pseudo}. 
In the case of human pose estimation, pseudo-labels come in the form of 2D heatmaps (i.e., \emph{pseudo-heatmaps}, Figure \ref{fig:pseudo_heatmap_overview}).
Here, the use of SSL is made significantly more challenging by the difficulty of predicting heatmaps that are reliable across {\it all} keypoints, as shown in Figure~\ref{fig:teaser}. 

To address these limitations we propose a novel training scheme for semi-supervised human pose estimation. Our method consists of several key designs, which significantly improve the quality of the heatmaps used as pseudo-labels, yielding state-of-the-art performance. In details, our method employs a dual-student framework~\cite{ke2019dual} and enhances it with two major contributions: a denoising scheme to improve the quality of each student's pseudo-heatmaps and an uncertainty-guided pseudo-heatmap selection to determine the best possible pseudo-heatmaps to use as supervision to train the students.

Our \emph{denoising} scheme improves the pseudo-heatmaps estimation in two ways: it ensembles the
outputs of multiple strong and weak augmented views to obtain better estimates of each joint location and it refines the actual responses at these locations using a threshold-and-refine scheme. 
Furthermore, to deal with erroneous high confidence predictions from poorly calibrated models \cite{nguyen2015deep}, we propose an uncertainty-based cross-student pseudo-heatmap \emph{selection}. Inspired by recent works in semi-supervised classification~\cite{rizve2021defense, NEURIPS2020_f23d125d} and object detection~\cite{xu2021end, liu2022unbiased}, our method discards noisy pseudo-labels by computing an approximate prediction uncertainty. 
%
Our solution differs from \cite{rizve2021defense, xu2021end, liu2022unbiased} in the way it parameterizes the uncertainty: instead of operating over simple one-hot vectors (classification) or bounding boxes coordinates (detection), we operate over full 2D heatmaps, one for each joint of a person's skeleton. To solve this complex problem, we propose to estimate the per-pixel uncertainty across all spatial locations. To the best of our knowledge we are the first to explore this direction for SSL human pose estimation.  



To evaluate our design, we experiment using two versions of MS COCO~\cite{lin2014microsoft}: \textit{COCO-Partial} and \textit{COCO-Additional}. Under both protocols (especially the former) our method achieves state-of-the-art performance and outperforms existing methods. For example, with 0.5K labeled instances, we outperform the strong DualPose~\cite{xie2021empirical} by 7.22 and 7.28 AP points (+25\% absolute improvement) when evaluated with a single model and a model ensemble respectively, and consistently outperforms DualPose by 4-5 AP points with 1K and 2K labeled instances (+12\%). With stronger input augmentation, we still consistently achieves more than 2 AP improvement over the best competitor. Finally, we conduct an extensive ablation study of the components of our design and validate their importance. 
\section{Related Work}
\label{sec:related}

\myparagraph{Semi-supervised learning}. Recent advances in semi-supervised learning have been achieved through various deep learning methods \cite{chen2022semi}, with self-training~\cite{sohn2020fixmatch, lee2013pseudo, arazo2020pseudo, bachman2014learning, rizve2021defense, iscen2019label} and consistency regularization~\cite{tarvainen2017mean, laine2016temporal, berthelot2019mixmatch, berthelot2020remixmatch, sajjadi2016regularization, xie2020unsupervised} being the most commonly used approaches. Self-training uses the model's own predictions to supervise the model itself. One well-known technique is pseudo-labelling~\cite{lee2013pseudo}, which converts model predictions to one-hot pseudo-labels and uses them to learn from unlabeled data. Consistency regularization~\cite{tarvainen2017mean, laine2016temporal, berthelot2019mixmatch, berthelot2020remixmatch, sajjadi2016regularization, xie2020unsupervised}, on the other hand, enforces consistent predictions across input or model perturbations. State-of-the-art SSL methods usually combine these two prominent techniques~\cite{sohn2020fixmatch, zhang2021flexmatch}. For example, FixMatch \cite{sohn2020fixmatch} uses weakly augmented views to generate pseudo-labels which are used as targets on the strongly augmented views to ensure consistently regularized output. Although this weak-strong data augmentation paradigm works well on image classification~\cite{sohn2020fixmatch, xie2020unsupervised} and object detection~\cite{liu2021unbiased, zhou2021instant, xu2021end}, weakly augmented views does not always give reliable learning targets for human pose estimation. This motivates us to propose more advanced formulations that generate more reliable pseudo-heatmaps for semi-supervised learning.  

\myparagraph{Semi-supervised human pose estimation}. Unlike research in semi-supervised 3D human pose estimation~\cite{mitra2020multiview, kanaujia2007semi, pavllo20193d}, research in semi-supervised 2D human pose estimation just starts to attract more interest recently \cite{ke2019dual}. Due to the natural difference between 3D and 2D tasks, those 3D semi-supervised methods cannot be directly transferred to the 2D scenarios. Early work~\cite{ukita2018semi} on label-efficient 2D human pose estimation combines semi-supervised and weakly-supervised schemes. Several works develop semi-supervised methods for key-point localization~\cite{honari2018improving, moskvyak2021semi, wang2022pseudo}. For instance, Moskvyak et al~\cite{moskvyak2021semi} uses semantic consistency constraints to regularize the network whereas PLACL~\cite{wang2022pseudo} combines curriculum learning with reinforcement learning to improve the training. Recently, DualPose~\cite{xie2021empirical} establishes a benchmark for semi-supervised human pose estimation and presents the dual student framework~\cite{ke2019dual, blum1998combining} based on the weak-strong data augmentation paradigm~\cite{sohn2020fixmatch, xie2020unsupervised}. In this work, we focus on developing semi-supervised methods for 2D human pose estimation, but our method can also be adapted to other key-point localization tasks.

\myparagraph{Uncertainty estimation in semi-supervised learning (SSL)}. Several research efforts have been made on the uncertainty estimation in network predictions~\cite{blundell2015weight, gal2016dropout, lakshminarayanan2017simple, malinin2018predictive, van2020uncertainty}. In the SSL domain, \cite{rizve2021defense} use Monte Carlo Dropout and measure uncertainty by calculating standard deviation of output probabilities, which helps to choose a better subset of pseudo-labels for classification. \cite{yu2019uncertainty} use comparable techniques but employ predictive entropy in the context of medical segmentation tasks, while uncertainty is used as weights to compute pseudo-labels in \cite{xia20203d}. In object detection, \cite{xu2021end} samples jittered box to measure localization uncertainty as the box regression variance, meanwhile~\cite{liu2022unbiased} explicitly model the box boundary uncertainty via an extra auxiliary branch. Our work differs from these existing works in two ways. First, we propose to  estimate uncertainty using pixel-level Gaussian heatmap regression. Second, our 
estimator is computationally efficient -- no new parameters are required for learning, which makes it scalable to estimate per-pixel uncertainty across all spatial locations.

\section{Methodology}
\label{sec:method}

\begin{figure*}[!t]
    \centering
    \includegraphics[width=\linewidth]{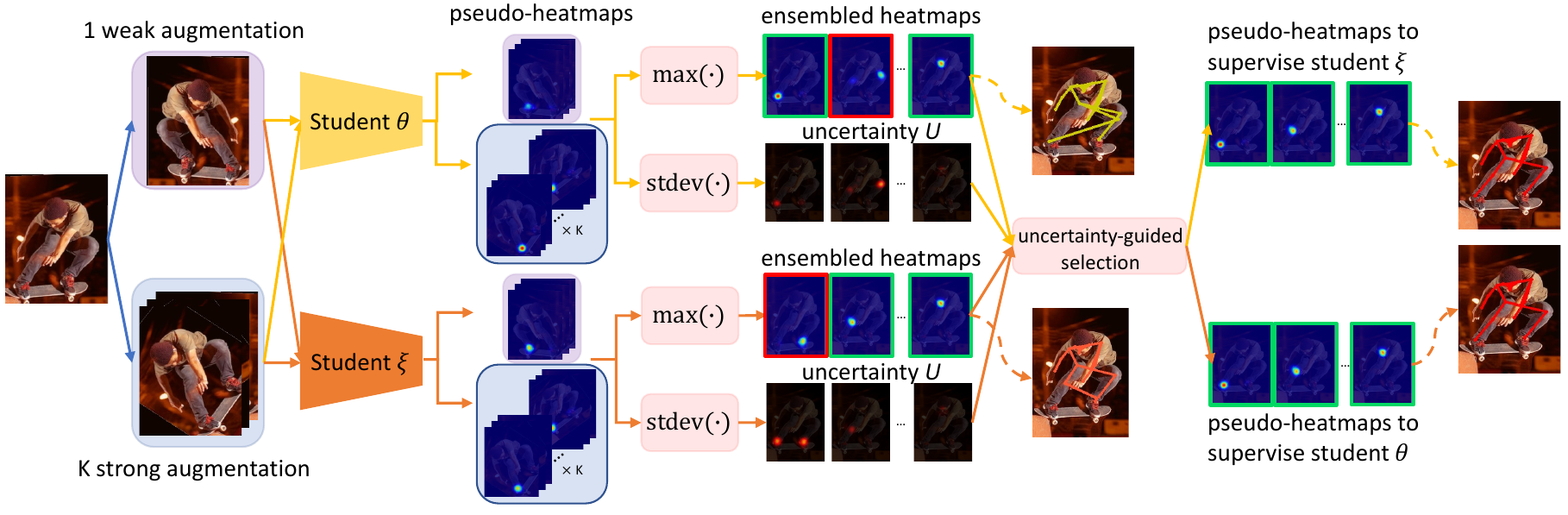}
    \caption{\small \it
    \textbf{ Our proposed semi-supervised pose estimation model.} 
    Given an input image, we apply multi-view augmentation to generate a set of pseudo heatmaps, and ensemble these pseudo-heatmaps to obtain more precise pseudo groundtruth from each student network (Sec \ref{sec:method_pla}). To select the better pseudo groundtruth among two student networks as learning targets on unlabeled data, we introduce an uncertainty estimator to estimate the uncertainty of each pseudo heatmap and guide the selection of pseudo heatmaps (Sec \ref{sec:method_ucms}).
    }
    \label{fig:pseudo_heatmap_overview}
    \vskip -1em
\end{figure*}


{As Figure \ref{fig:pseudo_heatmap_overview} shows, given an input image (either labeled or unlabeled), our semi-supervised pose estimator augments the image into different views, and generates pseudo-heatmaps based on multiple augmented views. The pseudo groundtruth of human pose is further produced guided by uncertainty. In the following, we first define our task and revisit a dual-student framework (Sec \ref{sec:basics}). We detail our approach which aggregates multiple augmented views to calibrate the pseudo-heatmaps (Sec \ref{sec:method_pla}), and select the more reliable learning targets guided by uncertainty (Sec \ref{sec:method_ucms}). }




{\subsection{Problem Definition and Preliminaries}}
\label{sec:basics}

\myparagraph{Problem Definition.}
{We consider the problem of semi-supervised human pose estimation. Given a small set of annotated images $ \mathcal{D}_l = \{ (x_i, y_i) \}_{i=1}^M$ (where $x_i$ is the image labeled with pose annotations $y_i$) and a large set of unlabeled images $\mathcal{D}_u = \{u_i\}_{i=1}^N$ (where $u_i$ is the unlabeled image), our goal is to develop a semi-supervised learning (SSL) framework that trains an human pose estimator with $\mathcal{D}_l$ and $\mathcal{D}_u$ jointly to achieve better model generalization.} 

{
A typical supervised pose estimator \cite{sohn2020simple, newell2016stacked, tompson2014joint} (with model parameters $\theta$) is trained to predict a set of heatmaps $H_i=\{h_{i,j}\}_{j=1}^J$ of $J$ different human joints for an input image $x_i$ by optimizing the following mean square errors: 
\begin{equation}
\begin{aligned}
\mathcal{L}^S_\theta &= || y_i - H_i ||^2 
\end{aligned}
\label{eq:mse}
\end{equation}
where $\mathcal{L}^S_\theta$ is the supervised loss for training network $\theta$. The key success of semi-supervised learning is to enable learning from unlabeled data with unsupervised loss terms. We detail how to formulate these loss terms in the following.
}



\myparagraph{Revisit the dual student framework.} 
{Following existing SSL methods which are often built upon consistency regularization \cite{tarvainen2017mean, laine2016temporal, berthelot2019mixmatch, berthelot2020remixmatch, sajjadi2016regularization, xie2020unsupervised}, a recent work introduces a dual student framework \cite{ke2019dual, xie2021empirical} for semi-supervised pose estimation, known as DualPose \cite{xie2021empirical}. DualPose trains two student networks $\theta$ and $\xi$ jointly, and employs the output from one network as the learning targets to supervise the other network. Specifically, an input image $I$ is processed with weak-strong augmentation \cite{sohn2020fixmatch, xie2020unsupervised} through affine transformations to obtain its weakly and strongly augmented counterparts $I_w, I_s$. Each student network predicts a set of heatmaps for the augmented inputs $I_w, I_s$:
\begin{equation}
\begin{aligned}
    H_{\theta, w} = \theta({I}_w) 
    \quad &\text{and} \quad 
    H_{\theta, s} = \theta({I}_s), \\
    H_{\xi, w} = \xi({I}_w) 
    \quad &\text{and} \quad 
    H_{\xi, s} = \xi({I}_s), 
\end{aligned}
\label{eq:dualpose1}
\end{equation}
where $\theta(\cdot), \xi(\cdot)$ are two student networks. 
$H_{\theta, w}, H_{\xi, w} \in \mathbb{R}^{J \times h \times w}$ are predictions on the weakly augmented view $I_w$ (where $J$ is the number of joints, $h\times w$ denotes the spatial dimensions of a heatmap). $H_{\theta, s}, H_{\xi, s} \in \mathbb{R}^{J \times h \times w}$ are predictions on the strongly augmented view $I_s$.}

{To derive the learning targets on unlabeled images, the heatmaps predicted on weakly augmented views are used as the pseudo ground-truth (i.e., \emph{pseudo-heatmaps}). The unsupervised loss terms are then computed as follows:
\begin{equation}
\begin{aligned}
\mathcal{L}^U_\theta &= || H_{\xi, w}-H_{\theta, s}||^2 = || H_{\xi, w}-\theta({I}_s)||^2 \\
\mathcal{L}^U_\xi &= || H_{\theta, w}-H_{\xi, s}||^2 = || H_{\theta, w}-\xi({I}_s)||^2  
\end{aligned}
\label{eq:dualpose2}
\end{equation}
where $\mathcal{L}^U_\theta, \mathcal{L}^U_\xi$ denote unsupervised loss terms for student networks $\theta, \xi$. $H_{\xi, w}, H_{\theta, w}$ are used as the learning targets that supervise the networks to learn from unlabeled data.
}

\noindent \emph{Limitations.} While being simple, this framework can produce unreliable learning targets on unlabeled data. Its pseudo-heatmaps can be noisy, as they are estimated from a single random weakly augmented view (Figure \ref{fig:teaser} left). As a consequence, the model may fit on this noise and exacerbate the training error propagation. In the next two sections we propose two novel formulations to address these problems.

\subsection{Denoising pseudo-heatmaps}
\label{sec:method_pla}

 Using the outputs on weakly augmented views as learning targets for strongly augmented views was shown to be effective for semi-supervised learning in image classification~\cite{sohn2020fixmatch, xie2020unsupervised, liu2021unbiased, zhou2021instant}. However, in semi-supervised pose estimation, we find that the pseudo-heatmaps produced with one weakly augmented view are sometimes inaccurate (especially on some more challenging joints) and tend to have low responses (see Figure \ref{fig:teaser} top). To mitigate this problem, we propose two denoising solutions: first, we ensemble the outputs of multiple strong and weak augmented views to obtain better estimates of each joint location; and second, we refine the actual responses at these locations. We present the details of these solutions next. 

\myparagraph{Multi-view augmentation.}
To improve the pseudo-heatmaps estimation, we apply $K$ different strong affine transformations on the input images to produce more variants for the same input image $\{{I}_{{s}, k}\}_{k=1}^K$. Our key insight is that by adding richer stochastic perturbations in the input space, we can cancel out mistakes and noises in learning targets caused by the randomness in one single weakly augmented view. This shares the same spirit as model ensembling \cite{mitchell1982generalization,breiman2001random}, where a committee of models
is used to cover different regions of the version space and improve the accuracy of the final predictions.
{Given multiple strongly augmented views $\{{I}_{{s}, k}\}_{k=1}^K$ of the input $I$, we obtain a set of candidate pseudo-heatmaps $\{H_{\theta, s}^k\}_{k=1}^K$, which can be further ensembled to derive more reliable learning targets. Following a winner-take-all strategy, we compute the ensembled pseudo-heatmaps by taking the maximum scores per pixel over outputs of all augmented views, i.e., $\{H_{\theta, s}^k\}_{k=1}^K$ and $H_{\theta, w}$. That is, the ensembled pseudo-heatmaps from the two student networks are 
\begin{equation}
\label{eq:targets}
\begin{aligned}
    P_{\theta} = \text{max}\{H_{\theta, w}, \{H_{\theta, s}^k \}_{k=1}^K \ \}, \\
    P_{\xi} = \text{max}\{H_{\xi, w}, \{H_{\xi, s}^k\}_{k=1}^K \ \}, 
\end{aligned}
\end{equation}
where $P_{\theta}, P_{\xi}$ are the pseudo-heatmaps obtained from networks $\theta, \xi$, which are calibrated with better accuracy thanks to ensembling multiple outputs of $K{+}1$ augmented views.} 


\myparagraph{A threshold-and-refine scheme.}
We employ a ``threshold-and-refine'' scheme to further denoise the pseudo-heatmaps $P_{\theta}, P_{\xi}$ in Eq. \eqref{eq:targets} in cleaner responses. 
This scheme is similar to the confidence-based thresholding scheme used to generate pseudo labels in existing semi-supervised image classification methods such as Pseudo-Label \cite{lee2013pseudo} and FixMatch \cite{sohn2020fixmatch}, but we specially design it for generating pseudo-heatmaps in human pose estimation. 
Specifically, if the maximum response of pseudo-heatmaps $P_{\theta}$ is above a pre-defined threshold $\tau$, we refine $P_{\theta}$ by applying a 2D Gaussian centered at the location corresponding to the maximum response of $P_{\theta}$, where the 2D Gaussian is a special operation used in human pose estimation to translate ground truth points into 2D heatmaps.

{With our threshold-and-refine scheme, we can derive refined pseudo-heatmaps $P_{\theta}, P_{\xi}$ to serve as more accurate unsupervised targets for training two student networks:
\begin{equation}
\begin{aligned}
\mathcal{L}^U_\theta &= || P_{\xi}-H_{\theta, s}||^2 = || P_{\xi}-\theta({I}_s)||^2, \\
\mathcal{L}^U_\xi &= || P_{\theta}-H_{\xi, s}||^2 = || P_{\theta}-\xi({I}_s)||^2,  
\end{aligned}
\label{eq:calibrate}
\end{equation}
where the above unsupervised loss terms constrain the output $\theta({I}_s), \xi({I}_s)$ of one strongly augmented view ${I}_s$. For both student networks, $P_{\theta}, P_{\xi}$ in Eq. \eqref{eq:calibrate} are more accurate pseudo-heatmaps compared to the heatmaps $H_{\theta, w}, H_{\xi, w}$ in Eq. \eqref{eq:dualpose2} generated from one weakly augmented views.} 

\subsection{Uncertainty-guided pseudo-heatmaps selection}
\label{sec:method_ucms}

The two student networks $\theta$ and $\xi$ can perform differently, even when predicting the same joint in the same image. Importantly, one student may output more reliable learning targets than the other student and we can exploit this information to further improve the final pseudo-heatmaps used for cross-training. 
For this, we propose to select the most accurate pseudo-hatmaps using an uncertainty-guided selection scheme that rejects the heatmaps with higher uncertainty scores. 

\myparagraph{Uncertainty estimation on heatmaps.}
{Given an unlabeled image ${I}$ and a set of heatmaps $\{H_{\theta, w}, \{H_{\theta, s}^k \}_{k=1}^K \ \}$ predicted on its augmented views by student network $\theta$, we can estimate the uncertainty of the heatmap for each single joint. Rather than following the prior work~\cite{rizve2021defense} that uses Monte Carlo Dropout~\cite{gal2016dropout} for uncertainty estimation, we propose a novel uncertainty estimator which computes the uncertainty based on the outputs of multiple augmented views. Specifically, the uncertainty is estimated as the pixel-wise standard deviation across the set of heatmaps predicted on multiple augmented views for each joint. The uncertainty can be expressed as follows for student networks $\theta, \xi$:
\begin{equation}
\begin{aligned}
    {U}_{\theta} &= \text{stdev}(\{H_{\theta, w}, \{H_{\theta, s}^k \}_{k=1}^K \ \}), \\
    {U}_{\xi} &= \text{stdev}(\{H_{\xi, w}, \{H_{\xi, s}^k \}_{k=1}^K \ \}),
\label{eq:uncertainty}
\end{aligned}
\end{equation}
where ${U}_{\theta}, {U}_{\xi} \in \mathbb{R}^{J \times h \times w}$ have the same dimensions as $H_{\theta, w}$, $H_{\theta, s}^k$, $H_{\xi, w}$, $H_{\xi, s}^k$, e.g., ${U}_{\theta}{=}\{U_{j}\}_{j=1}^J$ is a set of uncertainty maps for $J$ different joints. To further derive a scalar value that represents the uncertainty of each heatmap, we take the maximum value on uncertainty maps, which leads to a set of $J$ uncertainty scores: $u_{\theta} {=} \{u_{j}\}_{j=1}^J {=} \{\text{max}(U_{j})\}_{j=1}^J$, where $\text{max}(U_{j})$ is the maximum uncertainty value of the $j_{th}$ joint.
}


\myparagraph{Uncertainty-guided selection across student networks.} 
{Given the two set of uncertainty maps ${U}_{\theta}, {U}_{\xi}$ which measure the reliability of the two set of heatmaps produced by two student networks, we can select the more reliable outputs as pseudo-heatmaps for the unlabeled images. Specifically, we trust one student's own output over the other student when its uncertainty scores are lower than a certain margin $\Delta$, than those of the other student. The learning targets selected by uncertainty for student network $\theta$ are:
\begin{equation}
\label{eq:selection1}
\begin{aligned}
\hat{P}_{\theta,j} = \begin{cases}
{P}_{\theta,j}, & \text{if } \ u_{\theta,j} + \Delta < u_{\xi,j} \\
\mathcal{P}_{\xi,j}, & \text{otherwise}
\end{cases}
\end{aligned}
\end{equation}
where $j \in [1, J]$ denotes the index of $J$ different joints. $\hat{P}_{\theta,j} $ is the final pseudo-heatmap of the $j_{th}$ joint which is selected between ${P}_{\theta,j}$ and ${P}_{\xi,j}$ based on their uncertainty score $u_{\theta,j}$ and $u_{\xi,j}$. If the uncertainty $u_{\theta,j}$ of ${P}_{\theta,j}$ is lower than the uncertainty $u_{\xi,j}$ of ${P}_{\xi,j}$ than a certain margin $\Delta$, then ${P}_{\theta,j}$ is selected as pseudo-heatmap for the $j_{th}$ joint; otherwise ${P}_{\xi,j}$ is selected.} We compute the learning targets $\hat{P}_{\xi,j}$ for student network $\xi$ in an equivalent way as Eq. \eqref{eq:selection1}.
%
Finally, from these we can obtain more reliable pseudo-heatmaps $\hat{P}_{\theta}, \hat{P}_{\xi} \in \mathbb{R}^{J \times h \times w}$ for $J$ different joints to serve as the final pseudo-groundtruth for the student networks $\theta$ and $\xi$.

\myparagraph{Semi-supervised learning objective.}
With $P_{\theta}, P_{\xi}$ as the learning targets, we can now formulate the unsupervised loss on unlabeled data on two student networks $\theta$, $\xi$, in the similar spirit as Eq. \eqref{eq:calibrate}:
\begin{equation}
\label{eq:unsupervised_loss}
\begin{aligned}
\mathcal{L}_{\theta}^U = || H_{\theta, s} - \hat{P}_{\xi} ||^2, \quad
\mathcal{L}_{\xi}^U = || H_{\xi, s} - \hat{P}_{\theta} ||^2,
\end{aligned}
\end{equation}
where $\hat{P}_{\theta}, \hat{P}_{\xi}$ are the targets selected guided by uncertainty as in Eq. \eqref{eq:selection1}. The final semi-supervised learning objective is to jointly optimize the above unsupervised loss along with the supervised loss in Eq. \eqref{eq:mse}: 
\begin{equation}
\label{eq:semi_loss}
\begin{aligned}
\mathcal{L}_{\theta} = \mathcal{L}_{\theta}^S + \mathcal{L}_{\theta}^U, \quad
\mathcal{L}_{\xi} = \mathcal{L}_{\xi}^S + \mathcal{L}_{\xi}^U,
\end{aligned}
\end{equation}
where $\mathcal{L}_{\theta}^S, \mathcal{L}_{\xi}^S$ are the supervised loss computed on the labeled images based on Eq. \eqref{eq:mse}. $\mathcal{L}_{\theta}^U, \mathcal{L}_{\xi}^U$ are the unsupervised loss computed on unlabeled images based on Eq. \eqref{eq:unsupervised_loss}.

\section{Experiments}
\label{sec:experiments}

\subsection{Datasets and Evaluation}
\label{exp:data_eval}

\myparagraph{COCO Dataset.} COCO is a large-scale benchmark dataset for object detection, segmentation and human pose estimation. For semi-supervised human pose estimation, we use the following sets. {\bf train2017} contains 118K images and 150K labeled person images with 17 keypoints; {\bf unlabeled2017} contains 123K unlabeled images; {\bf val2017} contains 5K images and 6K labeled person images; {\bf test-dev} set contains 20K images where annotations are private (we obtain results by submitting our predictions to their evaluation server). Unless explicitly stated, we evaluate the models on the val2017. We validate the performance of our method under two evaluation protocols: \textit{COCO-Partial} and \textit{COCO-Additional}. \textit{COCO-Partial} splits train2017 into 0.5K, 1K, 2K, 5K, 10K incremental labeled sets and uses the remaining images as unlabeled.  \textit{COCO-Additional} uses the whole train2017 as labeled data and unlabeled2017 as unlabeled. 

\myparagraph{AI Challenger Dataset.} 
To demonstrate the effectivness of our method in leveraging additional unlabeled data, we also experiment with the large-scale {AI Challenger Dataset}~\cite{wu2019large}, consisting of 210k images. While these images come with annotations, we disregard them as consider them unlabelled.

\myparagraph{Evaluation metrics.} We report mean average precision (AP) over 10 Object Keypoint Similarity (OKS)~\cite{lin2014microsoft}'s thresholds: [0.5, 0.55 ..., 0.9, 0.95]. OKS is calculated as the Euclidean distances between each corresponding ground truth and the detected keypoints, normalized by the scale of the person. To measures each keypoint's localization accuracy, we also report PCK (Percentage of Correct Keypoints)~\cite{6380498} score. A joint is correct is it falls within $\alpha l$ pixels of the ground-truth position, where $\alpha$ is a constant and $l$ is maximum side length of the ground-truth person bounding box. The PCK@0.1 ($\alpha$ = 0.1) score is reported.



\subsection{Implementation details}
\myparagraph{Benchmark settings.} 
For fair comparison, we implement all models (baseline and ours) in the same codebase and compare them using the same backbones. On \textit{COCO-Partial}, we evaluate each method with 3 random seeds and report the mean and standard deviation across the 3 runs. We follow \cite{xie2021empirical} and use Simple Baseline~\cite{xiao2018simple} or HRNet~\cite{sun2019deep} as the pose estimator and conduct our experiments with various backbones. Specifically, for \textit{COCO-Partial}, we use ResNet18 \cite{he2016deep} for the low-data regimes (0.5K, 1K and 2K) and use ResNet50 for high-data regimes (5K and 10K). Since performance on ResNet18 gets saturated with more labeled data, we also use ResNet50, ResNet101, ResNet152 and HRNetW48 for evaluation. 


\begin{table*}[t]
\begin{center}
\resizebox{\linewidth}{!}{%
\begin{tabular}{llclllll}
\toprule
\textbf{Method}                   &\textbf{Aug}            & \multicolumn{1}{c}{\textbf{0.5K}}    & \multicolumn{1}{c}{\textbf{1K}}     & \multicolumn{1}{c}{\textbf{2K}}        & \multicolumn{1}{c}{\textbf{5K}}         & \multicolumn{1}{c}{\textbf{10K}}           \\
 \cmidrule(r){1-1}\cmidrule(lr){2-2}\cmidrule(lr){3-5}\cmidrule(lr){6-7} 
& & \multicolumn{3}{c}{\textbf{ResNet18}} & \multicolumn{2}{c}{\textbf{ResNet50}} \\
 \cmidrule(r){1-1}\cmidrule(lr){2-2}\cmidrule(lr){3-5}\cmidrule(lr){6-7}
Supervised Baseline      &A            & 22.05 $\pm$  1.12 & 30.91 $\pm$ 0.64 & 36.07 $\pm$ 0.50 & 49.31 $\pm$ 0.44 & 55.98 $\pm$ 0.08 \\
 \cmidrule(r){1-1}\cmidrule(lr){2-2}\cmidrule(lr){3-5}\cmidrule(lr){6-7}
DataDistill \cite{radosavovic2018data} $\dagger$             &A            & -               & 37.6    & -       & 51.6           & 56.6           \\
DualPose \cite{xie2021empirical}    &A            & 32.16 $\pm$ 1.18  & 41.54 $\pm$ 0.66 & 46.48 $\pm$ 0.44 & 58.39 $\pm$ 0.53 & 63.07 $\pm$ 0.47   \\
\bf \ProposedMethodName{}                 & A            & \textbf{39.38 $\pm$ 0.94~\scriptsize{\textcolor{red}{(+7.22)}}}       & \textbf{46.27 $\pm$ 0.50~\scriptsize{\textcolor{red}{(+4.73)}}}  &   \textbf{50.74 $\pm$ 0.28~\scriptsize{\textcolor{red}{(+4.26)}}}     & \textbf{60.67 $\pm$ 0.12~\scriptsize{\textcolor{red}{(+2.28)}}}       & \textbf{63.81 $\pm$ 0.32~\scriptsize{\textcolor{red}{(+0.74)}}}          \\ 

\cmidrule(r){1-1}\cmidrule(lr){2-2}\cmidrule(lr){3-5}\cmidrule(lr){6-7}
DualPose-Ensemble \cite{xie2021empirical} & A            & 32.98 $\pm$ 1.27  & 42.67 $\pm$ 0.67 & 48.19 $\pm$ 0.45 & 59.37 $\pm$ 0.44 & 64.46 $\pm$ 0.46   \\
\bf \ProposedMethodName{}-Ensemble        &A            & \textbf{40.26 $\pm$ 0.87~\scriptsize{\textcolor{red}{(+7.28)}}}           & \textbf{47.32 $\pm$ 0.70~\scriptsize{\textcolor{red}{(+4.65)}}}   &  \textbf{51.96 $\pm$ 0.35~\scriptsize{\textcolor{red}{(+3.77)}}}     & \textbf{61.62 $\pm$ 0.17~\scriptsize{\textcolor{red}{(+2.25)}}}          & \textbf{64.84 $\pm$ 0.25~\scriptsize{\textcolor{red}{(+0.38)}}}          \\  \cmidrule(r){1-1}\cmidrule(lr){2-2}\cmidrule(lr){3-5}\cmidrule(lr){6-7}

DualPose \cite{xie2021empirical}   & A+JC            & 36.89 $\pm$ 0.20  & 44.97 $\pm$ 0.14 & 48.67 $\pm$ 0.16 & 60.62 $\pm$ 0.25 & 64.25 $\pm$ 0.37   \\
\bf \ProposedMethodName{}           & A+JC            & \textbf{42.13 $\pm$ 0.21~\scriptsize{\textcolor{red}{(+5.24)}}}           & \textbf{47.58 $\pm$ 0.29~\scriptsize{\textcolor{red}{(+2.61)}}} &    \textbf{51.25 $\pm$ 0.27~\scriptsize{\textcolor{red}{(+2.58)}}}     & \textbf{62.14 $\pm$ 0.10~\scriptsize{\textcolor{red}{(+1.52)}}}          & \textbf{65.36 $\pm$ 0.15~\scriptsize{\textcolor{red}{(+1.11)}}}          \\  \cmidrule(r){1-1}\cmidrule(lr){2-2}\cmidrule(lr){3-5}\cmidrule(lr){6-7}

DualPose Ensemble \cite{xie2021empirical} & A+JC            & 37.56 $\pm$ 0.47  & 46.19 $\pm$ 0.14 & 50.48 $\pm$ 0.26 & 62.04 $\pm$ 0.17 & 65.69 $\pm$ 0.21   \\
\bf \ProposedMethodName{}-Ensemble       &A+JC            & \textbf{43.06 $\pm$ 0.20}~\scriptsize{\textcolor{red}{(+5.50)}}          & \textbf{48.70 $\pm$ 0.09~\scriptsize{\textcolor{red}{(+2.51)}}}  &   \textbf{52.54 $\pm$ 0.32~\scriptsize{\textcolor{red}{(+2.06)}}}     & \textbf{63.34 $\pm$ 0.02~\scriptsize{\textcolor{red}{(+1.30)}}}          & \textbf{66.46 $\pm$ 0.15~\scriptsize{\textcolor{red}{(+0.77)}}}          \\
\bottomrule
\end{tabular}
}
\end{center}
\vspace{-3mm}
\caption{\small \it \textbf{Comparison on \textit{COCO-Partial}}. 
We report results using different number of labeled images (i.e., 0.5K, 1K, 2K, 5K, 10K), and test the models under different augmentation strategies: `A' means affine transformation; `JC' denotes JointCutout from prior work~\cite{xie2021empirical}. 
Metric: AP. 
The best result in each setup is in {\bf bold}. 
We show the improved margins over the best competitor in \textcolor{red}{red}. 
}
\label{tab:coco-partial}
\vspace{-4mm}
\end{table*}

\myparagraph{Training details.}
Our backbone network is initialized by pre-training on ImageNet \cite{deng2009imagenet}. The base learning rate is 0.001 and Adam~\cite{kingma2014adam} optimizer is used. For \textit{COCO-Partial}, we train our model for 100 epochs and decay learning rate by a factor of 10 at the 70-th and 90-th epoch. For \textit{COCO-Additional} we train our model for 400 epochs and decay learning rate at the 300-th and 350-th epoch. 

\myparagraph{Data Augmentation.}
Our \emph{weak} augmentation policy\footnote{We keep our weak augmentations the same as DualPose~\cite{xie2021empirical} to ensure that our comparison is fair and can highlight the contribution of our model in selecting more reliable pseudo heatmaps for semi-supervised learning.} uses random affine (A) transformations with rotation degrees sampled from $[-30^{\circ}, 30^{\circ}]$ and scale factors sampled from $[0.75, 1.25]$. 
Our \emph{strong} augmentation policy samples rotation degrees from $[-60^{\circ}, 60^{\circ}]$ and scale factors from $[0.5, 1.5]$ with Beta distribution ($\alpha=\beta=0.75$).
Finally, when stated, we also train with JointCutout (JC)~\cite{xie2021empirical}. 

\myparagraph{Testing details.}
We apply a two-stage top-down paradigm similar to \cite{chen2018cascaded, papandreou2017towards}. On COCO val2017, we run the pose estimators on ground truth bounding boxes without image flipping. On COCO test-dev set, we use the popular person detector of Simple Baseline~\cite{xiao2018simple} (AP of 60.9). We follow previous works~\cite{chen2018cascaded, newell2016stacked} and predict joint locations as the average between the original and flipped images. 

\subsection{Comparison with the state-of-the-art}


\myparagraph{Comparison on \textit{COCO-Partial protocol}.}  
 We compare three semi-supervised pose estimation models: DataDistill \cite{radosavovic2018data}, DualPose~\cite{xie2021empirical}, and ours on COCO val2017, in terms of OKS-based AP. 
Table \ref{tab:coco-partial} demonstrates the stronger performance of our model and its ability to leverage more reliable pseudo-heatmaps to achieve better generalization:
our model consistently outperforms the best competitor DualPose, under all setups, including using different number of labeled images  (ranging from 0.5K to 10K), different backbone networks (ResNet18 and ResNet50), and different types of data augmentation strategies (A, and A+JC). 
In particular, we find it performs especially well in the low-label regime. For example, when using 0.5K or 1K labeled images, our improved margins over DualPose are significantly larger, up to +7.22 when using a simple affine (A) transformation (32.16 vs 39.38). 
Considering the high complexity and prohibitive cost of annotating human pose estimation datasets, we believe this low-regime to be the most interesting target for SSL, as it would enable training budget-friendly real-world pose estimation models on only few hundred samples, while still achieving very competitive performance. 

%

\noindent {\it Evaluating individual joint predictions' quality}. 
OKS-based AP computes aggregated statistics over multiple joints and a person's skeleton is considered correct if \emph{most} of the predicted joints falls within a certain distance from their corresponding ground truth (i.e., a single joint mistake is irrelevant for OKS when the majority are correct). To truly understand the improvement that our model brings over DualPose, we now investigate the per-joint performance using the Percentage of Correct Keypoint (PCK)  metric. 
%
%
As shown in Table \ref{tab:coco-partial-per-joint}, we achieve substantially better joint estimation performance across all 7 (meta-)joints (we average symmetric joints - left and right). Interestingly, the largest improvements can be observed on the challenging and dynamic limb joints (e.g., +8.1 on Wrist), showing the importance of modelling uncertainty. 

\begin{table}[!t]
\begin{center}
\resizebox{\linewidth}{!}{%
    \begin{tabular}{lccccccc}
            \toprule
            \textbf{Method} & \textbf{Ankl} & \textbf{Knee} & \textbf{Hip} & \textbf{Wrist} & \textbf{Elb} & \textbf{Shld} & \textbf{Head} \\

            \hline
            DualPose~\cite{xie2021empirical} & 61.4 & 62.8 & 62.5 & 62.1 & 68.9 & 77.2 & 91.7 \\
            \bf \ProposedMethodName{} & \bf65.7 & \bf67.6 & \bf68.6 & \bf70.2 & \bf75.8 & \bf81.7 & \bf93.5 \\
            \bottomrule
    \end{tabular}
}
\vspace{-5mm}
\end{center}
    \caption{\small \it
    \textbf{Comparison of per joint results on \textit{COCO-Partial}.} 
    Metric: PCK, which measures per-joint localization accuracy.
    }
    \label{tab:coco-partial-per-joint}
\vspace{-5mm}
\end{table}

\begin{table*}[!tbp]
    \centering
    \small
    \resizebox{0.9\linewidth}{!}{%
    \begin{tabular}{lcccclccccc}
        \toprule
        \textbf{Method} & \textbf{Unlabeled Data} & \textbf{Backbone} & \multicolumn{1}{c}{\textbf{AP}} & \multicolumn{1}{c}{$\bf{AP_{50}}$} & \multicolumn{1}{c}{$\bf{AP_{75}}$} & \multicolumn{1}{c}{$\bf{AP_{M}}$} & \multicolumn{1}{c}{$\bf{AP_{L}}$} & \multicolumn{1}{c}{\textbf{AR}} \\
        \hline
        SimpleBaseline \cite{xiao2018simple}    &   -   &   \multirow{4}{*}{ResNet50}    &   70.2    &   90.9    &   78.3  &    67.1    &    75.9    & 75.8 \\
        SB w/ DualPose~\cite{xie2021empirical}    &    COCO-unlabeled2017    &    &  72.3    & 91.8    & 80.5    & 69.3    & 77.8    & 77.7    \\
        SB w/ \ProposedMethodName{}    &    COCO-unlabeled2017    &     & 72.5   &  91.8   &  81.0  &  69.7   &  77.9   &  77.8   \\
        SB w/ \ProposedMethodName{}    &    COCO-unlabeled2017 + AIC    &   & \textbf{73.3}       &  \textbf{92.1}   &  \textbf{82.0}  &  \textbf{70.9}   &  \textbf{78.5}   &  \textbf{78.9}   \\
        \midrule
        SimpleBaseline \cite{xiao2018simple}    &   -   &   \multirow{4}{*}{ResNet152}   &   71.9    &   91.4    &   80.1 &    68.9    &    77.4    & 77.5 \\
        SB w/ DualPose~\cite{xie2021empirical}    &  COCO-unlabeled2017  &       & 73.7    & 92.1    & 82.1    & 71.0                             &    79.0    & 79.1    \\
        SB w/ \ProposedMethodName{}    &  COCO-unlabeled2017  &        & 73.8     &  91.9   &  82.1  &  71.1   &  79.2   &  79.2  \\
        SB w/ \ProposedMethodName{}    &  COCO-unlabeled2017 + AIC  &       & \textbf{74.2}    &   \textbf{92.1}  &  \textbf{82.4}  &  \textbf{71.5}   &  \textbf{79.6}   &  \textbf{79.4}  \\
        \bottomrule
    \end{tabular}    }
    \caption{\small \it
    \textbf{Comparison on \textit{COCO-Additional} with the test-dev set.} 
    The entire COCO training set (train2017) is used as labeled sets. COCO-unlabeled2017 set (and AI Challenger (AIC)~\cite{wu2019large}) is used as unlabeled set. The person detection results are provided by Simple Baseline (SB) \cite{xiao2018simple} and flipping strategy is used. \vspace{-2mm}}
    \label{table:cocotest}
    \vspace{-2mm}
\end{table*}

\myparagraph{Comparison on \textit{COCO-Additional protocol}.} 
To study the effect of training on datasets of larger scale, \textit{COCO-Additional} uses COCO train2017 as labeled set and COCO unlabeled2017 as unlabeled set, leading to a total of 118K labeled and 123k unlabeled images. To exploit unlabeled data in the wild, we also experiment by expanding the aforementioned unlabeled set with the {AI Challenger (AIC) Dataset}~\cite{wu2019large}, resulting in a total of 333K unlabeled images. We experiment with different backbone networks (ResNet50, ResNet152) {and compare with state-of-the-art competitors in Table \ref{table:cocotest}. We find that training with unlabeled data from COCO-unlabeled2017 and AI Challenger gives the best results for all backbones. For instance, when using ResNet50 as backbone, our model achieves 73.3 AP which is better than 72.3 by DualPose. }

\subsection{Ablation Study}
\label{sec:ablation}


In this section, we now ablate the components of our model. {Unless specified,} we use the \textit{COCO-Partial} protocol with 1K labeled images and evaluate on COCO val2017. We first provide model ablation in Section \ref{sec:ablation}, and then analyze individual components in Section \ref{sec:component1} and  \ref{sec:component2}. 

\begin{table}
	\begin{center}
	\resizebox{0.9\linewidth}{!}{

	    \begin{tabular}{ l | c  | c  c  c }
    		\toprule
    		 
    		 & DualPose & \multicolumn{3}{c}{\textbf{Ours}}\\
    		  \hline 
    		 Multi-View Augmentation &  & \cmark & \cmark & \cmark \\
    		 Threshold-and-Refine & & & \cmark & \cmark \\
    		 Uncertainty Guided Selection & & & & \cmark \\
    		 \hline 
    		 AP & 42.67 & 44.91 & 46.49 & \textbf{47.97} \\ 
    		 \bottomrule
    	\end{tabular}
	}
	\end{center}
	\vspace{-3mm}
	\caption{\small \it \textbf{Ablation study of different model components.}}
	\vspace{-5.5mm}
	\label{tab:ablation_model_component_table}
\end{table}



\subsubsection{Model ablation study}
\label{sec:ablation}

{\myparagraph{Different model components.} In Table~\ref{tab:ablation_model_component_table}, we evaluate different components of our approach and ablate how each of them contributes to the model performance. Building upon DualPose, we find that multi-view augmentation improves the performance by 2.24 points. Denoising with threshold-and-refine brings additional 1.58 AP. Further applying the uncertainty-guided pseudo-heatmaps selection, the performance reaches 47.97 AP, which is 5.3 points better than DualPose. This suggests the collective effects of different model components.}

{\myparagraph{Different backbones and more unlabeled data.} In Table~\ref{tab:coco-additional}, we train our model using varying amounts of unlabeled data (i.e.,  COCO unlabeled and COCO unlabeled + AIC) and compare them with a fully-supervised model solely trained on COCO train2017. As Table~\ref{tab:coco-additional} shows, we find that our approach outperforms the supervised baseline using all backbones, achieving a noteworthy increase of up to +3.2 AP when only using COCO unlabeled as unlabeled set. By using the additional AIC unlabeled dataset, our method further improves the performance by an additional margin, up to +1.4 AP. This shows that our model is capable of taking advantage of additional unlabelled data and further improve the performance of human pose estimation.}

\subsubsection{Analysis of denoising pseudo-heatmaps}
\label{sec:component1}

\myparagraph{Effect of multi-view augmentation.} As described in Sec \ref{sec:method_pla}, we augment each image into K strong and 1 weak views. 
In Table \ref{tab:ablation_augmentation} we evaluate this choice against baselines using only 1, $K$ and $K+1$ weak views. We also compare two ways of aggregating the $K+1$ pseudo-heatmaps obtained from augmented views: average and maximum response. 
Results show that taking the maximum response is better than average. Furthermore, adding K additional augmentation (either weak or strong) always greatly improves the model performance. This shows the benefits of introducing multi-view augmentation, as ensembling outputs from multiple views tends to cancel out the individual errors in single view, providing more accurate pseudo-heatmaps. Lastly, we find that using K strong views is more effective than using K weak ones, which improves AP from 43.74 to 44.91 (+1.17). This is because strong augmentations provide more diverse variations in input space, leading to more precise ensembled pseudo-heatmaps. 

\begin{table}[!t]
\begin{center}
\resizebox{\linewidth}{!}{%
    \begin{tabular}{lcccc}
            \toprule
            \textbf{Method} & \textbf{Backbone} & \textbf{Unlabeled Data} & $\bf{AP}$ & $\bf{AP_{75}}$ \\
            \hline
            Supervised & \multirow{3}{*}{ResNet50}  & - & 70.9 & 78.2 \\
            \bf \ProposedMethodName{}   &   & COCO-unlabeled2017 & 74.1 & 81.5\\
            \bf \ProposedMethodName{}   &   & COCO-unlabeled2017+AIC & \textbf{75.5} & \textbf{82.7} \\
            \hline
            Supervised & \multirow{3}{*}{ResNet101}     & - & 72.5 & 80.3 \\
            \bf \ProposedMethodName{}  &  & COCO-unlabeled2017 & 75.7 & 83.6 \\
            \bf \ProposedMethodName{}   &   & COCO-unlabeled2017+AIC & \textbf{76.6} & \textbf{84.6} \\
            \hline
            Supervised & \multirow{3}{*}{ResNet152} & - &      73.2 & 81.2 \\
            \bf \ProposedMethodName{}  &  & COCO-unlabeled2017 & 76.0 & 83.7\\
            \bf \ProposedMethodName{}   &   & COCO-unlabeled2017+AIC & \textbf{76.7} & \textbf{84.6} \\ 
            \hline
            Supervised & \multirow{3}{*}{HRNetW48}      & - & 77.2        &    84.6    \\
            \bf \ProposedMethodName{}  &  & COCO-unlabeled2017 & 79.4 & 86.7 \\
            \bf \ProposedMethodName{}   &   & COCO-unlabeled2017+AIC & \textbf{79.8} & \textbf{86.9}  \\
            \bottomrule
    \end{tabular}
}
    \vspace{-5mm}
\end{center}
    \caption{\small \it \textbf{Ablation study of additional unlabeled data.} Using more unlabeled data consistently improves model performance.
    }
    \label{tab:coco-additional}
    \vspace{-4mm}
\end{table}

\begin{table}[!t]
\begin{center}
\resizebox{0.65\linewidth}{!}{%
\begin{tabular}{llll}
            \toprule
            Row & Augmentation & Aggregate & mAP  \\
            \midrule
            1 & 1 weak & N/A  & 42.67  \\ 
            \midrule
            2 & (1 + K) weak  & \texttt{avg} &  42.11     \\  
            3 &(1 + K) weak  & \texttt{max} &  43.74     \\
            \midrule
            4 & K weak + 1 strong  & \texttt{avg} & 41.77     \\  
            5 & K weak + 1 strong  & \texttt{max} & 43.92     \\
            \midrule
            6 & 1 weak + K strong & \texttt{avg} & 39.58  \\
            7 & 1 weak + K strong & \texttt{max} & \textbf{44.91}  \\
            \bottomrule
    \end{tabular}
}
\end{center}
    \vspace{-3mm}
    \caption{\small \it \textbf{Effect of the multi-view augmentation.}
    Row 1 is the baseline with 1 weak augmentation. 
    Row 2-3/4-5/6-7 compare models with K additional weak/strong augmentation. 
    Note: we can aggregate multi-view pseudo heatmaps by taking the maximum response (i.e., \texttt{max}, Eq. \eqref{eq:targets}) or average response (i.e., \texttt{avg}). \vspace{-2mm}
    }
    \label{tab:ablation_augmentation}
    \vspace{-6mm}
\end{table}


\begin{figure*}[!ht]
    \centering
    \includegraphics[width=0.8\linewidth]{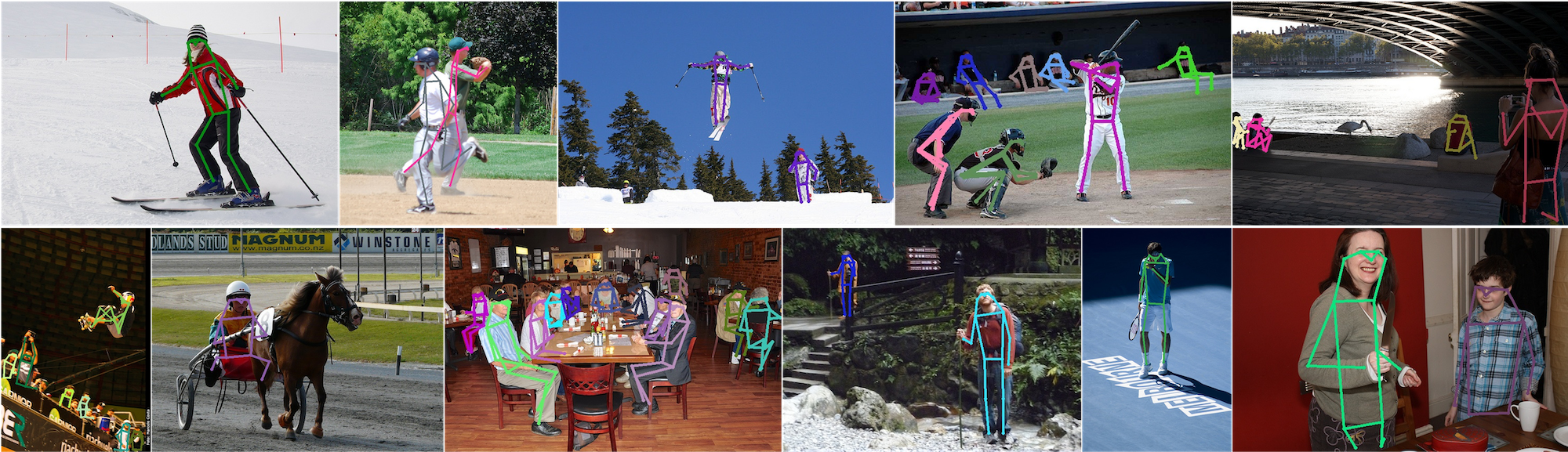}
    \vspace{-1mm}
    \caption{
        \small \it \textbf{Qualitative results} showing example images in the COCO datasets, which covers persons in different viewpoints, appearance changes, and performing different activities. These results show that our model estimates human pose of good quality on unlabeled data.
   }
   \label{fig:visual}
   \vspace{-7mm}
\end{figure*}

\begin{figure}[!t]
    \centering
    \includegraphics[width=0.75\columnwidth]{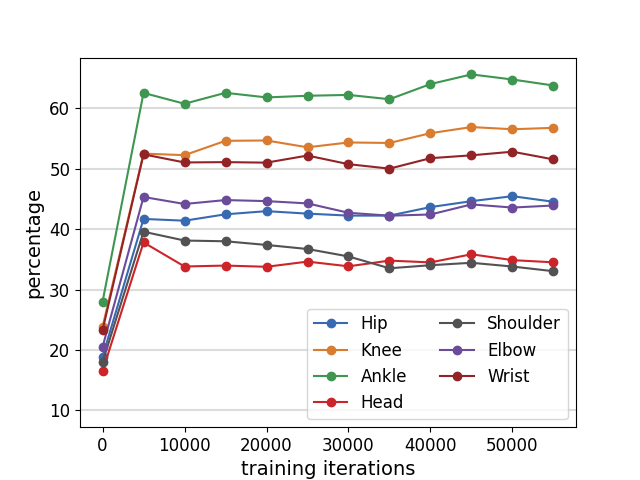}
    \caption{\small \it  \textbf{Analysis of pseudo-heatmaps selected from strongly augmented views}. The model tends to select pseudo-heatmaps from weakly-augmented views for torso joints, while selecting pseudo-heatmaps from strongly-augmented views for limb joints. 
    }
    \label{fig:strong_selection} 
    \vspace{-4mm}
\end{figure}

\myparagraph{Analysis of strong augmentation over different joints.} In Eq. \eqref{eq:targets} we obtain the maximum response heatmap across K strong and 1 weak augmented view. To understand the importance of strong views, we compute the percentage of pseudo-heatmaps selected from them for each joint independently (Figure \ref{fig:strong_selection}). Interestingly, there is no unanimous consensus and the percentage varies from $30$ (Head) to $60$ (Ankle), which explains the results of Table~\ref{tab:coco-partial-per-joint}: our method achieves similar performance as DualPose on Head ($+1.8$ PCK), as they both rely on weak augmentations to learn, but it improves error-prone limb joints by up to $+8.1$ PCK, thanks to its leveraging of strong augmentations. 


\begin{table}[!t]
    \centering
    \resizebox{0.8\linewidth}{!}{%
        \begin{tabular}{l|ccc}
            \toprule
            Row & Threshold & w/ Refinement & w/o Refinement  \\
            \hline
            1 & 0 & 44.32  & 43.91  \\
            2 & 0.1 & \textbf{47.97} & 44.73  \\
            3 & 0.2 & 45.80  & 45.07  \\
            4 &0.3  & 45.18 &  44.65     \\
            \bottomrule
        \end{tabular}
}
    \vskip 0.5em
    \caption{
        \small \it \textbf{Effect of the ``threshold-and-refine'' scheme.}
        From row 1 to 4, we increase the threshold when applying thresholding on pseudo heatmaps, and compare results with and without refining pseudo heatmaps with 2D Gaussion. Metric: AP.
    }
    \label{tab:ablation_threshold_and_reconstruct}
    \vspace{-6mm}
\end{table}

\myparagraph{Effect of ``threshold-and-refine'' scheme.} 
%
We evaluate this scheme in Table \ref{tab:ablation_threshold_and_reconstruct}. Our model achieves the lowest AP ($43.91$) when using neither thresholding nor refining. Moreover, refining pseudo heatmaps with 2D Gaussian always improves the model performance, especially with a threshold of $0.1$ (AP $47.97$, $+4.06$ over not using either). This shows the benefits of applying thresholding to remove the low response in pseudo heatmaps, and refining the heatmaps using 2D Gaussian to produce cleaner pseudo heatmaps.


\subsubsection{Analysis of uncertainty-guided selection}
\label{sec:component2}

\begin{figure}
    \centering
       \includegraphics[width=0.75\columnwidth]{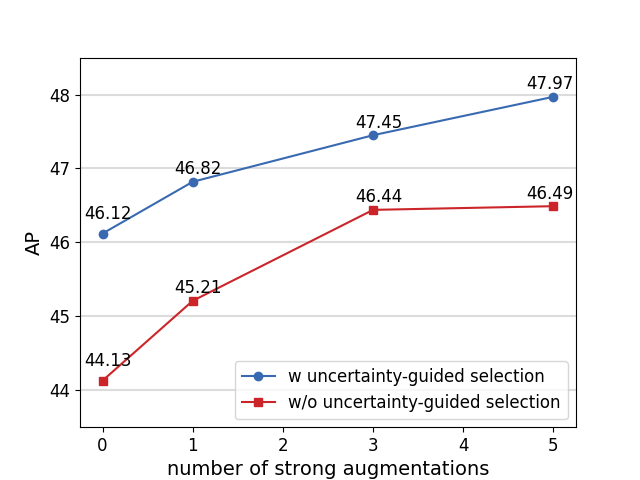}
    \caption{ \small \it 
    \textbf{Effect of uncertainty-guided pseudo heatmaps selection.} 
    We study our model with and without the uncertainty-guided selection 
    when using different number of augmented views.
    }
    \label{fig:each_component} 
    \vspace{-6mm}
\end{figure}

As mentioned in Sec \ref{sec:method_ucms}, we introduce an uncertainty estimator (Eq. \eqref{eq:uncertainty}) to estimate the uncertainty per joint based on a set of heatmaps from multiple augmented views. We also employ the uncertainty as an indicator to select more reliable pseudo groundtruth among two student networks to learn from unlabeled data (Eq. \eqref{eq:selection1}). To study their effect, we conduct an ablation study by removing the uncertainty-guided selection component from our model (Figure \ref{fig:each_component}).
We observe that using the uncertainty estimator to select the pseudo-heatmaps (blue curve) gives significant better performance than not using it (red curve). Interestingly, both curves (with and without uncertainty) improve their performance as we increase the number of augmented views, validating the previous hypothesis that multiple views are important and showcasing the complementarity of our multi-view and uncertainty contributions. 
\section{Conclusion}

We presented a new semi-supervised learning approach for human pose estimation. We introduceed multi-view augmentation to generate a candidate pool of pseudo heatmaps and selected the more reliable pseudo heatmaps guided by uncertainty. We exploited the pseudo-heatmaps as groundtruth on unlabeled data to perform semi-supervised learning. Our experimental results show that our model outperforms the state-of-the-art human pose estimator, especially in the extreme low-label regime where we have only a small fraction of labeled data (e.g., 0.5K, 1K). We also show that our model can effectively exploit unlabeled data in the wild to further boost its performance. 

{\small
\bibliographystyle{ieee_fullname}
\bibliography{reference}

\begin{thebibliography}{10}\itemsep=-1pt

\bibitem{arazo2020pseudo}
Eric Arazo, Diego Ortego, Paul Albert, Noel~E O’Connor, and Kevin McGuinness.
\newblock Pseudo-labeling and confirmation bias in deep semi-supervised learning.
\newblock In {\em IJCNN}, 2020.

\bibitem{bachman2014learning}
Philip Bachman, Ouais Alsharif, and Doina Precup.
\newblock Learning with pseudo-ensembles.
\newblock {\em arXiv preprint arXiv:1412.4864}, 2014.

\bibitem{berthelot2020remixmatch}
David Berthelot, Nicholas Carlini, Ekin~D Cubuk, Alex Kurakin, Kihyuk Sohn, Han Zhang, and Colin Raffel.
\newblock Remixmatch: Semi-supervised learning with distribution matching and augmentation anchoring.
\newblock In {\em ICLR}, 2020.

\bibitem{berthelot2019mixmatch}
David Berthelot, Nicholas Carlini, Ian Goodfellow, Nicolas Papernot, Avital Oliver, and Colin Raffel.
\newblock Mixmatch: A holistic approach to semi-supervised learning.
\newblock {\em arXiv preprint arXiv:1905.02249}, 2019.

\bibitem{blum1998combining}
Avrim Blum and Tom Mitchell.
\newblock Combining labeled and unlabeled data with co-training.
\newblock In {\em CLT}, 1998.

\bibitem{blundell2015weight}
Charles Blundell, Julien Cornebise, Koray Kavukcuoglu, and Daan Wierstra.
\newblock Weight uncertainty in neural network.
\newblock In {\em ICML}, 2015.

\bibitem{breiman2001random}
Leo Breiman.
\newblock Random forests.
\newblock {\em Machine learning}, 2001.

\bibitem{chapelle2009semi}
Olivier Chapelle, Bernhard Scholkopf, and Alexander Zien.
\newblock Semi-supervised learning (chapelle, o. et al., eds.; 2006)[book reviews].
\newblock {\em IEEE TNN}, 2009.

\bibitem{chen2022semi}
Yanbei Chen, Massimiliano Mancini, Xiatian Zhu, and Zeynep Akata.
\newblock Semi-supervised and unsupervised deep visual learning: A survey.
\newblock {\em IEEE TPAMI}, 2022.

\bibitem{chen2018cascaded}
Yilun Chen, Zhicheng Wang, Yuxiang Peng, Zhiqiang Zhang, Gang Yu, and Jian Sun.
\newblock Cascaded pyramid network for multi-person pose estimation.
\newblock In {\em CVPR}, 2018.

\bibitem{deng2009imagenet}
Jia Deng, Wei Dong, Richard Socher, Li-Jia Li, Kai Li, and Li Fei-Fei.
\newblock Imagenet: A large-scale hierarchical image database.
\newblock In {\em CVPR}, 2009.

\bibitem{gal2016dropout}
Yarin Gal and Zoubin Ghahramani.
\newblock Dropout as a bayesian approximation: Representing model uncertainty in deep learning.
\newblock In {\em ICML}, 2016.

\bibitem{he2016deep}
Kaiming He, Xiangyu Zhang, Shaoqing Ren, and Jian Sun.
\newblock Deep residual learning for image recognition.
\newblock In {\em CVPR}, 2016.

\bibitem{honari2018improving}
Sina Honari, Pavlo Molchanov, Stephen Tyree, Pascal Vincent, Christopher Pal, and Jan Kautz.
\newblock Improving landmark localization with semi-supervised learning.
\newblock In {\em CVPR}, 2018.

\bibitem{iscen2019label}
Ahmet Iscen, Giorgos Tolias, Yannis Avrithis, and Ondrej Chum.
\newblock Label propagation for deep semi-supervised learning.
\newblock In {\em CVPR}, 2019.

\bibitem{jeong2019consistency}
Jisoo Jeong, Seungeui Lee, Jeesoo Kim, and Nojun Kwak.
\newblock Consistency-based semi-supervised learning for object detection.
\newblock In {\em NeurIPS}, 2019.

\bibitem{kanaujia2007semi}
Atul Kanaujia, Cristian Sminchisescu, and Dimitris Metaxas.
\newblock Semi-supervised hierarchical models for 3d human pose reconstruction.
\newblock In {\em CVPR}, 2007.

\bibitem{ke2019dual}
Zhanghan Ke, Daoye Wang, Qiong Yan, Jimmy Ren, and Rynson~WH Lau.
\newblock Dual student: Breaking the limits of the teacher in semi-supervised learning.
\newblock In {\em ICCV}, 2019.

\bibitem{kingma2014adam}
Diederik~P Kingma and Jimmy Ba.
\newblock Adam: A method for stochastic optimization.
\newblock {\em arXiv preprint arXiv:1412.6980}, 2014.

\bibitem{laine2016temporal}
Samuli Laine and Timo Aila.
\newblock Temporal ensembling for semi-supervised learning.
\newblock {\em arXiv preprint arXiv:1610.02242}, 2016.

\bibitem{lakshminarayanan2017simple}
Balaji Lakshminarayanan, Alexander Pritzel, and Charles Blundell.
\newblock Simple and scalable predictive uncertainty estimation using deep ensembles.
\newblock {\em NeurIPS}, 30, 2017.

\bibitem{lee2013pseudo}
Dong-Hyun Lee et~al.
\newblock Pseudo-label: The simple and efficient semi-supervised learning method for deep neural networks.
\newblock In {\em ICMLW}, 2013.

\bibitem{lin2014microsoft}
Tsung-Yi Lin, Michael Maire, Serge Belongie, James Hays, Pietro Perona, Deva Ramanan, Piotr Doll{\'a}r, and C~Lawrence Zitnick.
\newblock Microsoft coco: Common objects in context.
\newblock In {\em ECCV}, 2014.

\bibitem{liu2021unbiased}
Yen-Cheng Liu, Chih-Yao Ma, Zijian He, Chia-Wen Kuo, Kan Chen, Peizhao Zhang, Bichen Wu, Zsolt Kira, and Peter Vajda.
\newblock Unbiased teacher for semi-supervised object detection.
\newblock {\em arXiv preprint arXiv:2102.09480}, 2021.

\bibitem{liu2022unbiased}
Yen-Cheng Liu, Chih-Yao Ma, and Zsolt Kira.
\newblock Unbiased teacher v2: Semi-supervised object detection for anchor-free and anchor-based detectors.
\newblock In {\em CVPR}, 2022.

\bibitem{malinin2018predictive}
Andrey Malinin and Mark Gales.
\newblock Predictive uncertainty estimation via prior networks.
\newblock {\em NeurIPS}, 2018.

\bibitem{mitchell1982generalization}
Tom~M Mitchell.
\newblock Generalization as search.
\newblock {\em Artificial intelligence}, 1982.

\bibitem{mitra2020multiview}
Rahul Mitra, Nitesh~B Gundavarapu, Abhishek Sharma, and Arjun Jain.
\newblock Multiview-consistent semi-supervised learning for 3d human pose estimation.
\newblock In {\em CVPR}, 2020.

\bibitem{moskvyak2021semi}
Olga Moskvyak, Frederic Maire, Feras Dayoub, and Mahsa Baktashmotlagh.
\newblock Semi-supervised keypoint localization.
\newblock {\em arXiv preprint arXiv:2101.07988}, 2021.

\bibitem{NEURIPS2020_f23d125d}
Subhabrata Mukherjee and Ahmed Awadallah.
\newblock Uncertainty-aware self-training for few-shot text classification.
\newblock In H. Larochelle, M. Ranzato, R. Hadsell, M.F. Balcan, and H. Lin, editors, {\em Advances in Neural Information Processing Systems}, volume~33, pages 21199--21212. Curran Associates, Inc., 2020.

\bibitem{newell2016stacked}
Alejandro Newell, Kaiyu Yang, and Jia Deng.
\newblock Stacked hourglass networks for human pose estimation.
\newblock In {\em ECCV}, 2016.

\bibitem{nguyen2015deep}
Anh Nguyen, Jason Yosinski, and Jeff Clune.
\newblock Deep neural networks are easily fooled: High confidence predictions for unrecognizable images.
\newblock In {\em CVPR}, 2015.

\bibitem{papandreou2017towards}
George Papandreou, Tyler Zhu, Nori Kanazawa, Alexander Toshev, Jonathan Tompson, Chris Bregler, and Kevin Murphy.
\newblock Towards accurate multi-person pose estimation in the wild.
\newblock In {\em CVPR}, 2017.

\bibitem{pavllo20193d}
Dario Pavllo, Christoph Feichtenhofer, David Grangier, and Michael Auli.
\newblock 3d human pose estimation in video with temporal convolutions and semi-supervised training.
\newblock In {\em CVPR}, 2019.

\bibitem{radosavovic2018data}
Ilija Radosavovic, Piotr Doll{\'a}r, Ross Girshick, Georgia Gkioxari, and Kaiming He.
\newblock Data distillation: Towards omni-supervised learning.
\newblock In {\em CVPR}, 2018.

\bibitem{rizve2021defense}
Mamshad~Nayeem Rizve, Kevin Duarte, Yogesh~S Rawat, and Mubarak Shah.
\newblock In defense of pseudo-labeling: An uncertainty-aware pseudo-label selection framework for semi-supervised learning.
\newblock {\em arXiv preprint arXiv:2101.06329}, 2021.

\bibitem{sajjadi2016regularization}
Mehdi Sajjadi, Mehran Javanmardi, and Tolga Tasdizen.
\newblock Regularization with stochastic transformations and perturbations for deep semi-supervised learning.
\newblock {\em arXiv preprint arXiv:1606.04586}, 2016.

\bibitem{sohn2020fixmatch}
Kihyuk Sohn, David Berthelot, Nicholas Carlini, Zizhao Zhang, Han Zhang, Colin~A Raffel, Ekin~Dogus Cubuk, Alexey Kurakin, and Chun-Liang Li.
\newblock Fixmatch: Simplifying semi-supervised learning with consistency and confidence.
\newblock In {\em NeurIPS}, 2020.

\bibitem{sohn2020simple}
Kihyuk Sohn, Zizhao Zhang, Chun-Liang Li, Han Zhang, Chen-Yu Lee, and Tomas Pfister.
\newblock A simple semi-supervised learning framework for object detection.
\newblock {\em arXiv preprint arXiv:2005.04757}, 2020.

\bibitem{sun2019deep}
Ke Sun, Bin Xiao, Dong Liu, and Jingdong Wang.
\newblock Deep high-resolution representation learning for human pose estimation.
\newblock In {\em CVPR}, 2019.

\bibitem{tarvainen2017mean}
Antti Tarvainen and Harri Valpola.
\newblock Mean teachers are better role models: Weight-averaged consistency targets improve semi-supervised deep learning results.
\newblock In {\em NeurIPS}, 2017.

\bibitem{tompson2014joint}
Jonathan~J Tompson, Arjun Jain, Yann LeCun, and Christoph Bregler.
\newblock Joint training of a convolutional network and a graphical model for human pose estimation.
\newblock In {\em NeurIPS}, 2014.

\bibitem{ukita2018semi}
Norimichi Ukita and Yusuke Uematsu.
\newblock Semi-and weakly-supervised human pose estimation.
\newblock {\em CVIU}, 2018.

\bibitem{van2020uncertainty}
Joost Van~Amersfoort, Lewis Smith, Yee~Whye Teh, and Yarin Gal.
\newblock Uncertainty estimation using a single deep deterministic neural network.
\newblock In {\em ICML}. PMLR, 2020.

\bibitem{wang2022pseudo}
Can Wang, Sheng Jin, Yingda Guan, Wentao Liu, Chen Qian, Ping Luo, and Wanli Ouyang.
\newblock Pseudo-labeled auto-curriculum learning for semi-supervised keypoint localization.
\newblock {\em arXiv preprint arXiv:2201.08613}, 2022.

\bibitem{wu2019large}
Jiahong Wu, He Zheng, Bo Zhao, Yixin Li, Baoming Yan, Rui Liang, Wenjia Wang, Shipei Zhou, Guosen Lin, Yanwei Fu, et~al.
\newblock Large-scale datasets for going deeper in image understanding.
\newblock In {\em ICME}, 2019.

\bibitem{xia20203d}
Yingda Xia, Fengze Liu, Dong Yang, Jinzheng Cai, Lequan Yu, Zhuotun Zhu, Daguang Xu, Alan Yuille, and Holger Roth.
\newblock 3d semi-supervised learning with uncertainty-aware multi-view co-training.
\newblock In {\em WACV}, 2020.

\bibitem{xiao2018simple}
Bin Xiao, Haiping Wu, and Yichen Wei.
\newblock Simple baselines for human pose estimation and tracking.
\newblock In {\em ECCV}, 2018.

\bibitem{xie2020unsupervised}
Qizhe Xie, Zihang Dai, Eduard Hovy, Thang Luong, and Quoc Le.
\newblock Unsupervised data augmentation for consistency training.
\newblock In {\em NeurIPS}, 2020.

\bibitem{xie2020self}
Qizhe Xie, Minh-Thang Luong, Eduard Hovy, and Quoc~V Le.
\newblock Self-training with noisy student improves imagenet classification.
\newblock In {\em CVPR}, 2020.

\bibitem{xie2021empirical}
Rongchang Xie, Chunyu Wang, Wenjun Zeng, and Yizhou Wang.
\newblock An empirical study of the collapsing problem in semi-supervised 2d human pose estimation.
\newblock In {\em ICCV}, 2021.

\bibitem{xu2021end}
Mengde Xu, Zheng Zhang, Han Hu, Jianfeng Wang, Lijuan Wang, Fangyun Wei, Xiang Bai, and Zicheng Liu.
\newblock End-to-end semi-supervised object detection with soft teacher.
\newblock In {\em ICCV}, 2021.

\bibitem{6380498}
Yi Yang and Deva Ramanan.
\newblock Articulated human detection with flexible mixtures of parts.
\newblock {\em IEEE TPAMI}, 2013.

\bibitem{yu2019uncertainty}
Lequan Yu, Shujun Wang, Xiaomeng Li, Chi-Wing Fu, and Pheng-Ann Heng.
\newblock Uncertainty-aware self-ensembling model for semi-supervised 3d left atrium segmentation.
\newblock In {\em MICCAI}, 2019.

\bibitem{zhang2021flexmatch}
Bowen Zhang, Yidong Wang, Wenxin Hou, Hao Wu, Jindong Wang, Manabu Okumura, and Takahiro Shinozaki.
\newblock Flexmatch: Boosting semi-supervised learning with curriculum pseudo labeling.
\newblock In {\em NeurIPS}, 2021.

\bibitem{zhou2021instant}
Qiang Zhou, Chaohui Yu, Zhibin Wang, Qi Qian, and Hao Li.
\newblock Instant-teaching: An end-to-end semi-supervised object detection framework.
\newblock In {\em CVPR}, 2021.

\bibitem{zhu2009introduction}
Xiaojin Zhu and Andrew~B Goldberg.
\newblock Introduction to semi-supervised learning.
\newblock {\em Synthesis lectures on artificial intelligence and machine learning}, 2009.

\bibitem{zhu2005semi}
Xiaojin~Jerry Zhu.
\newblock Semi-supervised learning literature survey.
\newblock Technical report, University of Wisconsin-Madison Department of Computer Sciences, 2005.

\end{thebibliography}
}

\end{document}